\documentclass[a4paper]{article}

\usepackage{INTERSPEECH2019}
\usepackage{xcolor}
\usepackage{subcaption}
\usepackage[sort&compress,numbers,square]{natbib}
\usepackage{tabularx}
\usepackage{enumitem}
\newcommand{\heading}[1]{}
\newcommand{\comments}[1]{}
\newcolumntype{Y}{>{\centering\arraybackslash}X}
\definecolor{cardinal}{cmyk}{0,1,0.63,0.29}
\definecolor{cardinal}{rgb}{0,0,0}

\usepackage{refcount}
\usepackage[%
breaklinks=true,
colorlinks=true, 
citecolor=cardinal,
urlcolor=cardinal,
linkcolor=cardinal,
linktocpage,
draft=false,
pageanchor=true,
hyperfigures=true,
plainpages=false
]{hyperref}

\makeatletter
\renewcommand{\section}{\@startsection
  {section}%
  {1}%
  {}%
  {-0.35\baselineskip}%
  {0.25\baselineskip}%
  {}}%

\renewcommand{\subsection}{\@startsection
  {subsection}%
  {2}%
  {}%
  {-0.1\baselineskip}%
  {0.1\baselineskip}%
  {}}%

\renewcommand{\subsubsection}{\@startsection
  {subsubsection}%
  {3}%
  {}%
  {-\baselineskip}%
  {0.5\baselineskip}%
  {}}%
\makeatother

\def\sevenpt{\def\baselinestretch{0.89}\let\normalsize\footnotesize\normalsize}

\ninept

\title{Spoken Language Intent Detection using Confusion2Vec}
\name{Prashanth Gurunath Shivakumar$^*$\thanks{$^*$ The authors contributed equally to this work}, Mu Yang$^*$, Panayiotis Georgiou}
\address{Signal Processing for Communication Understanding and Behavior Analysis Laboratory\\
University of Southern California, Los Angeles, USA}
\email{pgurunat@usc.edu, yangmu@usc.edu, georgiou@sipi.usc.edu}

\begin{document}

\maketitle
\begin{abstract}
Decoding speaker's intent is a crucial part of spoken language understanding (SLU).
The presence of noise or errors in the text transcriptions, in real life scenarios make the task more challenging.
In this paper, we address the spoken language intent detection under noisy conditions imposed by automatic speech recognition (ASR) systems.
We propose to employ confusion2vec word feature representation to compensate for the errors made by ASR and to increase the robustness of the SLU system.
The confusion2vec, motivated from human speech production and perception, models acoustic relationships between words in addition to the semantic and syntactic relations of words in human language.
We hypothesize that ASR often makes errors relating to acoustically similar words, and the confusion2vec with inherent model of acoustic relationships between words is able to compensate for the errors.
We demonstrate through experiments on the ATIS benchmark dataset, the robustness of the proposed model to achieve state-of-the-art results under noisy ASR conditions.
Our system reduces classification error rate (CER) by 20.84\% and improves robustness by 37.48\% (lower CER degradation) relative to the previous state-of-the-art going from clean to noisy transcripts.
Improvements are also demonstrated when training the intent detection models on noisy transcripts.

\end{abstract}

\section{Introduction}
\heading{SLU}
Spoken Language Understanding (SLU) systems aim at extracting semantic information from human spoken utterances. Such systems play a significant role in practical applications like personal AI voice assistants (e.g. Alexa, Siri, etc.), phone-call routing, booking system and so on. A SLU system is typically modeled as two separate components: an ASR front-end, which translates acoustic signal into text, followed by a Natural Language Understanding (NLU) module that performs inference for downstream tasks. Typical tasks include Domain classification, Intent Detection and Slot filling. In this work, we focus on the SLU system that performs Intent Detection, a task identifying speaker’s intent from speech. Such task is usually treated as an utterance classification problem \cite{yaman2008integrative}.


\heading{Intent Detection}

\subsection{Prior Work}
\heading{Prior Work Intent Detection (NLU)}
In the light of success of Deep Learning techniques, applying Deep Neural Networks on intent detection has been shown to be effective, often outperforming conventional classifiers, such as Support Vector Machines \cite{haffner2003optimizing}. In recent years, the NLU community have applied various techniques to improve  intent detection performance on manual transcripts. \cite{xu2013convolutional, guo2014joint, zhang2016joint} jointly model intent detection with slot filling, simplifying the NLU task by a unified model. 
\cite{hakkani2016multi, kim2017onenet} extend the joint modeling with domain knowledge, which enables information from multiple tasks to benefit the individual tasks and allow the NLU model to be applied to multiple-domain tasks. Going one step further, \cite{liu2016joint, zhang2018joint} involve adapting domain-specific language model (LM) while performing intent detection and slot filling, improving the performance on both LM and language understanding task. \cite{Liu+2016} explores strategies in joint modeling intent classification and slot filling using explicit alignment information provided by slot filling using attention-based encoder-decoder structure. On the basis of attention-based model, \cite{goo2018slot} connects context information from intent detection with slot filling using a gate mechanism. \cite{li2018self} employs a similar intent-augmented gating mechanism to guide the learning of the slot filling task. It further incorporates character-level embedding along with word-level embedding achieving state-of-the-art results in intent detection.
\comments{https://www.ijcai.org/Proceedings/16/Papers/425.pdf\\}

\heading{Prior work intent detection (SLU)}
However, suffering from ASR front-end errors, such as mis-recognized words, insertions and deletions, the performance of such systems degrades significantly, as shown in \cite{he2003data, deoras2013deep, mesnil2015using} and is still the bottleneck in SLU systems. On one hand, in order to make system more ASR-robust, ASR hypotheses can be incorporated into the model's training corpus. \cite{kurata2012leveraging, tur2013semantic, Masumura2018NeuralCC} exploit Word Confusion Networks to efficiently connect NLU models with ASR hypotheses. \cite{Simonnet2018Simulating} simulated ASR errors by randomly substituting words with their linguistically and acoustically similar candidates. On the other hand, there have been works that aim to jointly perform NLU tasks and ASR error adaption. \cite{schumann2018incorporating, zhu2018robust} employ Recurrent Neural Network (RNN) based Encoder-Decoder structure to reconstruct correct utterances from ASR hypotheses while performing intent detection and slot filling. \cite{stehwienfirst} makes richer feature representations by adding acoustic pitch accent flags into word embedding. 

\begin{figure*}[t]
\centering
\resizebox{0.9\linewidth}{0.75\columnwidth}{%
\begin{subfigure}[t]{0.49\linewidth}
\centering
\includegraphics[width=\linewidth]{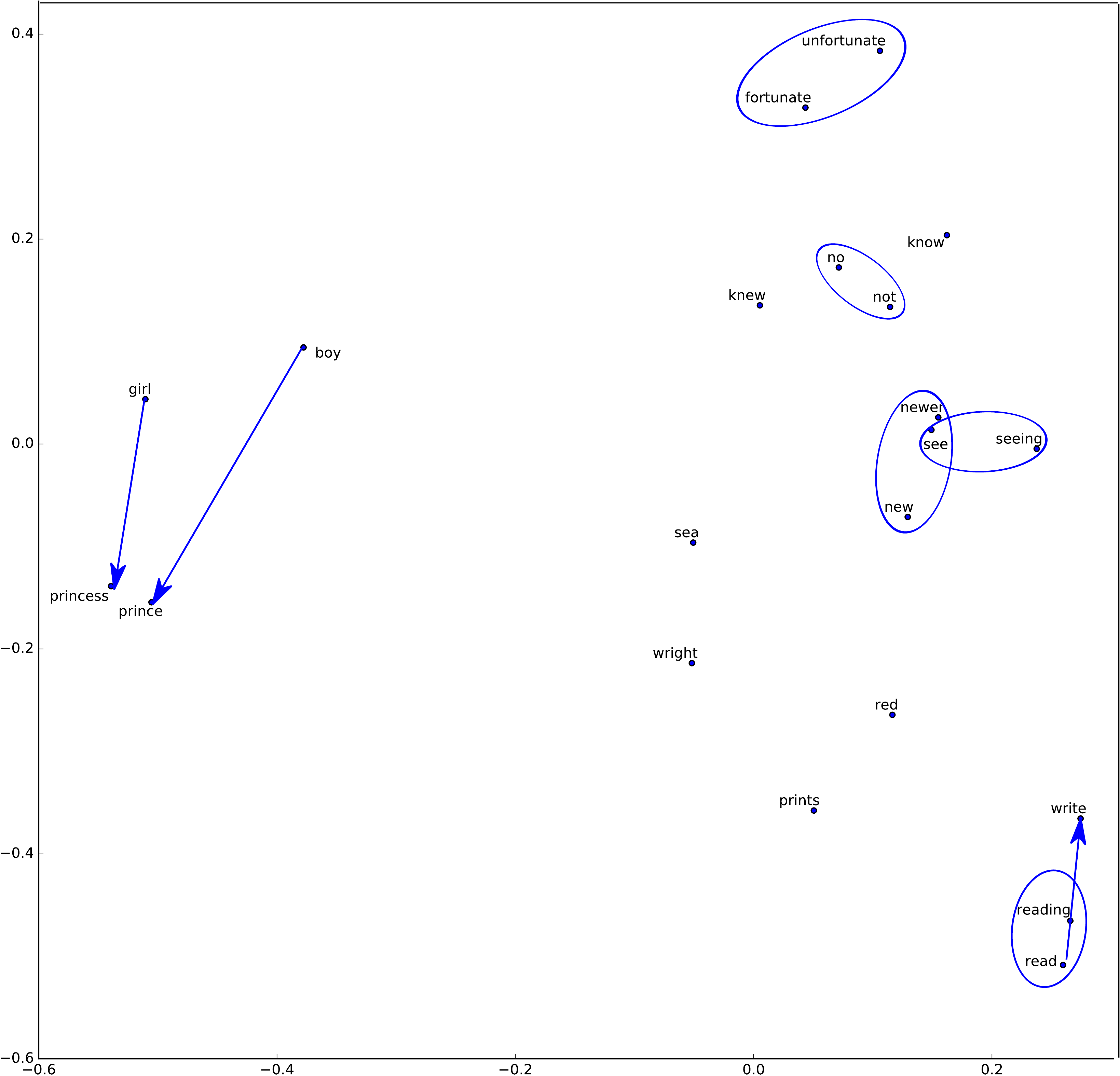}
\caption{Word2vec Space}\label{fig:w2v}
\end{subfigure}\hspace{15mm}
\begin{subfigure}[t]{0.47\linewidth}
\centering
\includegraphics[width=\linewidth]{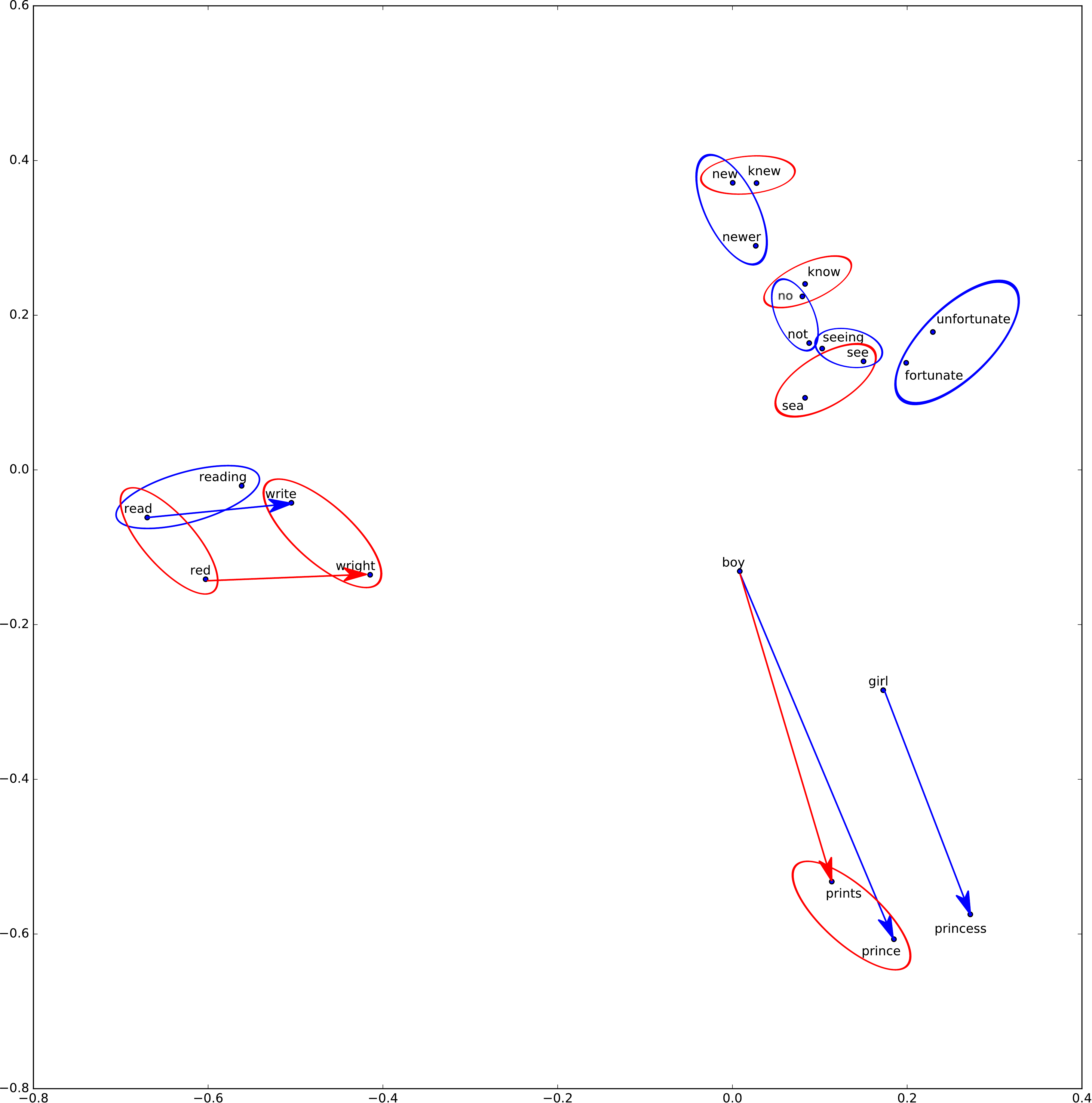}
\caption{Confusion2vec Space}\label{fig:c2v}
\end{subfigure}}
\captionsetup{justification=centering}
\caption{2D Vector space illustration after PCA dimension reduction for Word2vec and Confusion2vec\\
\footnotesize{The blue ellipses indicate syntactic word relations. The red ellipses indicate acoustic similarity relations.\\
The blue arrows illustrate the semantic relationships. The red arrows illustrate the interaction of acoustic similarity with semantic relationships.\\
The word2vec space is rich in semantic and syntactic word relations, however no trivial acoustic similarity is evident.\\
The confusion2vec space preserves the semantic and syntactic word relations, moreover captures additional acoustic similarity information.}
}\label{fig:word2vec_vs_conf2vec}
\vspace{-5mm}
\end{figure*}

\heading{Our Contribution}
In this paper, we specifically target the task of spoken language intent detection on noisy ASR transcripts.
In contrast to the majority of the works which mostly deal with the innovation of classification models \cite{liu2016joint, Liu+2016, goo2018slot, li2018self}, in our study, we concentrate on robust word feature representations.
We propose to employ the confusion2vec \cite{shivakumar2018confusion2vec} word vector representation to compensate for the errors made by an ASR and to provide enhanced and robust performance for the task of spoken language intent detection.
Confusion2vec captures acoustic similarity information of words in addition to the semantic-syntactic relations and is trained in a completely unsupervised manner on ASR lattices decoded on an out-of-domain corpora.
Moreover, unlike the studies which adapt the ASR to the target datasets and tasks \cite{schumann2018incorporating,liu2016joint}, we treat the ASR as a generic independent module, but contribute towards bridging the gap between the ASR and the NLU model.
We demonstrate with our experiments, on the benchmark ATIS dataset \cite{hemphill1990atis}, the vital role of the confusion2vec to the robustness of the intent classification.

\heading{Structure of the paper}
The rest of the paper is structured as follows.
In section~\ref{sec:methodology}, we present the proposed methodology, provide brief description of the confusion2vec word embedding and the intent classification model.
Section~\ref{sec:database_exp} describes the databases employed, our experimental setup and the baseline systems.
In section~\ref{sec:results}, we present and discuss the experimental results.
Finally, section~\ref{sec:conclusion} concludes the study and discusses the future work.

\section{Proposed Technique}\label{sec:methodology}
In this section, we first describe the confusion2vec word vector representation for the task of spoken language intent detection and then introduce the recurrent neural network intent classification model.

\subsection{Confusion2vec Word Embedding}

\heading{Motivation for Confusion2vec}
The role of word vector representations is crucial for NLP \cite{collobert2011natural}.
Efficient and information rich word embeddings like word2vec \cite{mikolov2013distributed}, glove \cite{pennington2014glove} are shown to capture semantics and syntactics of the language.
Using such efficient word representations have proven to be beneficial in the NLU tasks like named entity detection \cite{sienvcnik2015adapting}, intent detection \cite{hashemi2016query}.
The SLU tasks like intent detection \cite{kim2016intent,luan2015efficient,firdaus2018intent,balodis2018intent}, slot-filling \cite{firdaus2018deep}, spoken dialogue systems \cite{ferreira2015zero} have also benefited from using information rich word embeddings.
However, they are less than optimal in the cases of erroneous transcriptions, for example ASR transcriptions \cite{schumann2018incorporating,liu2016joint}, since the errors corrupt the semantic-syntactic space over local context of occurrence and thereby introduce noise in the model.

\heading{Confusion2vec}
In this work, we propose to employ recently proposed confusion2vec word vector representation \cite{shivakumar2018confusion2vec} for the task of intent detection to counter for errors present in the spoken transcriptions.
Motivated from human speech production and perception, the confusion2vec models the acoustic relations of words in addition to the semantic and syntactic relationships of words \cite{shivakumar2018confusion2vec}.
The confusion2vec uses unsupervised training techniques similar to skip-gram of word2vec, but operates on lattice-like structures or confusion networks output by the ASR.
Since the confusion networks of a typical ASR exhibits confusions between words on two principle axes (i) contextual, and (ii) acoustic similarity, the confusion2vec is devised to operate on both the axes, thereby modeling local context information (like word2vec) as well as acoustic similarity information.
Figure~\ref{fig:word2vec_vs_conf2vec} illustrates the 2-dimensional word vector space for word2vec and confusion2vec after dimension reduction using principal component analysis (PCA).
From the figure (and from extensive analysis done in \cite{shivakumar2018confusion2vec}), it is evident that confusion2vec space captures acoustic similarity between words without compromising the information captured by the word2vec.
Complex meaningful, useful interactions between the acoustic subspace and the semantic-syntactic subspaces are also observed.
For more information we would like to point the interested readers towards \cite{shivakumar2018confusion2vec}, which in detail describes and analyzes the confusion2vec embedding.

\heading{Application to intent detection}
In application to the spoken language intent detection task, the nature of ASR errors are often acoustically related.
Confusion2vec incorporates real, unsupervised, ASR output as its training corpus, thus the feature representation incorporates confusions (errors) nearby in its embedding space.
In other words, we hypothesize that the embedded acoustic similarity information in confusion2vec limits the impact of errors made by the ASR, and thus allows subsequent NLP tasks to be minimally affected.
We expect the following with respect to the intent detection task:
\begin{itemize}[wide, labelwidth=!, labelindent=0pt]
\item We expect our model to be less affected from ASR errors and thus achieve better performance in the case of noisy ASR transcriptions.
\item We expect our model to be at-least on par with word2vec under clean conditions.
\end{itemize}

\subsection{Intent Classification Model}\label{sec:rnn_arch}
\begin{figure}
\centering
\resizebox{0.8\columnwidth}{0.67\columnwidth}{%
\includegraphics[width=\linewidth]{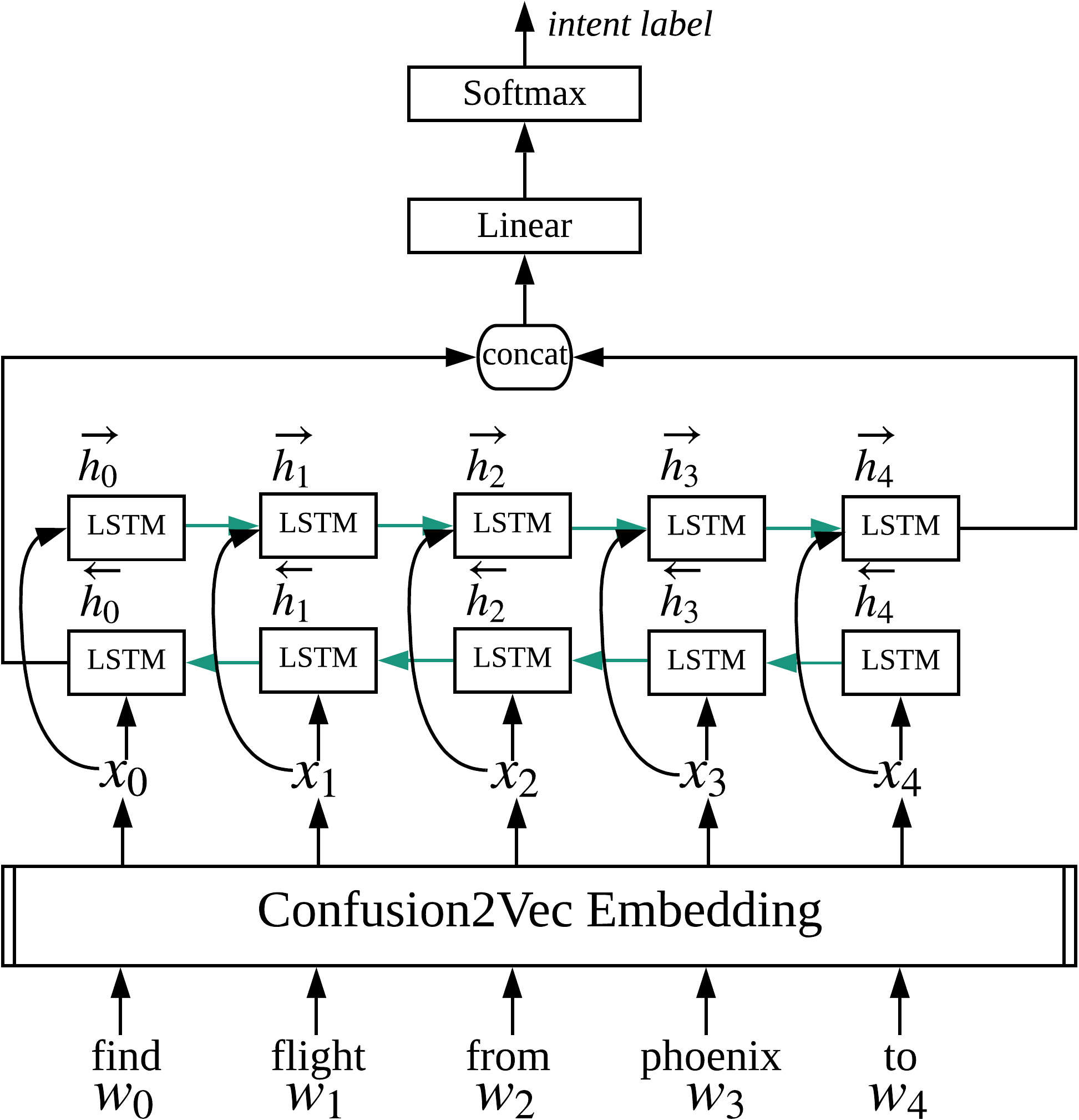}
\captionsetup{justification=centering}}
\caption{Intent Classification RNN Model}\label{fig:model_struct}
\vspace{-5mm}
\end{figure}

Since the contribution of this work is towards word feature representations, we employ a fairly simple recurrent neural network model for the classification task.
However, we believe the contributions on feature representations are orthogonal to the classification model and thus expect even better performance for more complex models like in \cite{liu2016joint,Liu+2016,goo2018slot,li2018self}. In this work, we use Bi-directional Long Short-Term Memory (LSTM) units, as shown in Figure~\ref{fig:model_struct}. Given an input utterance $w_0, w_1, ..., w_{T}$, each word in the input sequence is mapped to its word vector representation $x_0, x_1, ..., x_{T}$ by embedding look up. We formulate the model outputs as

{\footnotesize
\begin{align}
    \overrightarrow{h_t} = \overrightarrow{LSTM}(\overrightarrow{h_{t-1}}, x_t; \overrightarrow{\Theta}) 	&\qquad	 \overleftarrow{h_t} = \overleftarrow{LSTM}(\overleftarrow{h_{t+1}}, x_t; \overleftarrow{\Theta})
\end{align}
\begin{equation}
    \hat{P}_{\text{intent}} = Softmax(W~[\overrightarrow{h_T}, \overleftarrow{h_0}] + b)
\end{equation}
}%
where $h_t$ is the LSTM output of each direction at each time step $t$, $\Theta$ is the parameter of the LSTM. We feed the concatenation of two directional LSTM outputs at the last time step into the linear output layer (with weights $W$ and bias $b$) which projects it into the intent label space. Finally, the intent label is predicted from the Softmax-normalized probability distribution over all intent classes.

\section{Database \& Experimental Setup}\label{sec:database_exp}
\subsection{Database}
The ASR is trained on the Fisher English Training (LDC2004S13 and LDC2005S13) Speech corpora \cite{cieri2004fisher}.
The confusion2vec is trained on the output of the ASR, i.e., the confusion networks generated via Fisher English Corpora.
The database setup for the ASR and confusion2vec is identical and explained in detail in \cite{shivakumar2018confusion2vec}.

%

We trained the intent detection model on ATIS (Airline Travel Information Systems) dataset \cite{hemphill1990atis}, which comes with audio recordings and corresponding manual transcripts about humans asking for flight information. Following \cite{hakkani2016multi, goo2018slot}, we apply the same train, development, test split setup.
The setup contains 4478, 500 and 893 intent-labeled reference utterances in train, development and test set respectively. 
The database split and the corresponding audio mappings are made available to public\footnote{https://github.com/pgurunath/slu\_confusion2vec}.
In order to evaluate our model's robustness to ASR outputs, we also construct our ASR output set by decoding the corresponding audio recording for each of the data splits using the ASR.
In cases where an utterance is labeled with multiple intent labels, the top intent was selected as the true label, yielding 18 intents in total.


\subsection{Experimental Setup}\label{sec:exp_setup}
The training setup for the ASR and the confusion2vec is identical to our previous work \cite{shivakumar2018confusion2vec}.
For decoding the ATIS dataset, through our ASR, the audio samples were down-sampled from 16kHz to 8kHz.
The ASR achieves a WER rate of 18.54\% on the ATIS test set.
We choose the confusion2vec model yielding the best performance in \cite{shivakumar2018confusion2vec}, i.e., independently trained C2V-1 and C2V-c models are concatenated and jointly optimized with intra-Confusion2vec scheme (556 dimensions).

For intent detection, we train models on the 4478 utterances in training set, and tune hyper-parameters based on the classification accuracy on the 500 reference utterances in development set. The model with the best performance on the development set is chosen and evaluated on both reference test set and ASR test set. The hyper-parameter space we experimented with is as follows: Batch size is set to 1, i.e. each sentence is viewed as an independent sample. Hidden dimension of LSTM unit is tuned over $\{256, 128, 64, 32\}$, and dropout is tuned over $\{0.1, 0.2, 0.25\}$. We select Adam optimizer, with learning rate set to be among $\{0.001, 0.0005\}$. The maximum number of epochs is set to 50 with early-stopping strategy.

\subsection{Baseline Systems}

\heading{Scratch, W2V and GloVe}
The first set of baselines compare different conventional word embeddings.
They include:
(i) random initialization (556 dimensions) sampled from a uniform distribution,
(ii) vanilla GloVe\footnote{https://nlp.stanford.edu/projects/glove/} (300 dimensions) as in \cite{pennington2014glove}, and
(iii) skip-gram Word2Vec\footnote{\label{footnote:w2v}https://code.google.com/archive/p/word2vec} (556 dimensions) fine-tuned on Fisher English corpus reference transcripts (for fair comparison with confusion2vec). 
We also tried the vanilla Google word2vec\footnotemark[\getrefnumber{footnote:w2v}].
However, the performance was found to be consistently lower than the fine-tuned version, thus, we don't include it in the comparisons.
Note, only the randomly initialized word embedding is trainable, while all other embeddings are fixed throughout the training.
All the above baselines use identical RNN architecture for intent classification as described in section~\ref{sec:rnn_arch}.

\heading{Previous state-of-the-art models}
The second set of baselines compare our proposed model with the recent state-of-the-art models, including: (i) a joint intent detection, slot filling \& LM model \cite{liu2016joint}, (ii) an attention-based joint model that incorporates alignment information provided by slot filling task \cite{Liu+2016}, 
and (iii) an intent-augmented gating mechanism based model which further incorporates character-level embedding along with word-level embedding \cite{li2018self}.
The baselines are trained under our experiment setup using the same hyper-parameters reported in their original papers. 
We also reproduce the result of each model under their original experiment settings\footnotemark[\getrefnumber{footnote_exp_set}], and report the obtained scores in parentheses for reference.
For the adapted model trained on ASR, we consider a joint intent slot filling and intent detection model which performs sentence reconstruction from ASR hypotheses \cite{schumann2018incorporating} as the baseline, and report the scores on ASR outputs and corresponding ASR WER from original paper. 

\section{Results and Discussion}\label{sec:results}

\subsection{Training on Reference Clean Transcripts}

\begin{figure}[!hb]
\vspace{-2mm}
\includegraphics[width=\linewidth,trim={5mm 0 25mm 0},clip]{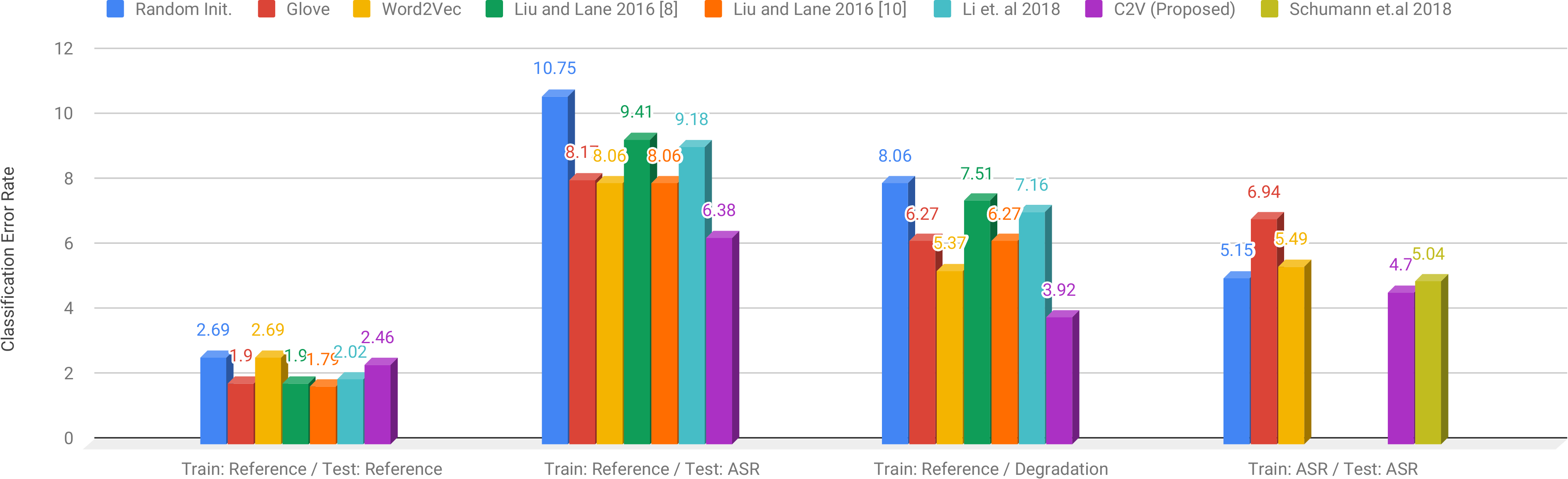}
\vspace{-5mm}
\captionsetup{justification=centering}
\caption{Comparison of CER for different systems}\label{fig:results}
\vspace{-2mm}
\end{figure}

Table~\ref{tab:results} and Figure~\ref{fig:results} illustrate the results obtained training on clean transcripts.
First, we compare the results between different word feature embedding (refer to upper half of Table~\ref{tab:results}).
On clean reference transcripts, GloVe embeddings provides the best performance (as found in \cite{firdaus2018intent}).
Both word2vec and random initialization provide identical results.
The proposed confusion2vec gives considerably lower CER compared to the Word2Vec and random initialization.
Although GloVe outperforms confusion2vec, we believe the comparison of confusion2vec is more fair with word2vec, since both use skip-gram modeling.
With the proposed Confusion2vec system, we don't expect improvements on clean reference transcripts, since the acoustic similarity/confusion is less relevant.
As expected, we observe no degradation in performance with confusion2vec and is on par with the popular, leading word vector representations for the task of intent detection on clean transcriptions.

\begin{table}[t]
  \centering
  \resizebox{\columnwidth}{!}{%
  \begin{tabularx}{1.15\columnwidth}{c *{3}{Y}}
    \toprule
      \textbf{Model}   & \textbf{Reference}    & \textbf{ASR}  & \textbf{$\Delta_{\text{diff}}$} \\
    \midrule
    Random                                  & 2.69          & 10.75         & 8.06 \\
    GloVe \cite{pennington2014glove}        & 1.90          & 8.17          & 6.27 \\
    Word2Vec \cite{mikolov2013distributed}  & 2.69          & 8.06         & 5.37 \\
    C2V (proposed)                          & 2.46          & \textbf{6.38} & \textbf{3.92} \\
    \midrule
    \citet{liu2016joint}                    & 1.90 (1.57)          & 9.41 (8.29)\footnotemark[\getrefnumber{footnote}]           & 7.51 (6.72) \\
    \citet{Liu+2016}                        & \textbf{1.79} (1.90)          & 8.06 (8.40)           & 6.27 (6.50) \\
    \citet{li2018self}                      & 2.02 (\textbf{1.34})          & 9.18 (9.07)          & 7.16 (7.73) \\
    \bottomrule
  \end{tabularx}}
  \captionsetup{justification=centering}
  \caption[caption for table]{Results with Training on Reference: Classification Error Rates (CER) for Reference and ASR Transcripts.\\
  \footnotesize{$\Delta_{\text{diff}}$ is the absolute degradation of model from clean to ASR.\\
    The numbers inside parenthesis indicate CER obtained reproducing the result of each model under their original experiment settings\footnotemark[\getrefnumber{footnote_exp_set}].}}\label{tab:results}
  \vspace{-8mm}
\end{table}

On noisy ASR transcripts, we see an increase in CER with all models.
Although, random initialization performed identical to Word2Vec on clean transcriptions, we see Word2Vec performs relatively better on ASR transcriptions.
This observation confirms that better word feature representations exhibit higher robustness to errors.
Similar trend is apparent with GloVe embeddings in comparison with random initialization, although we observe slightly higher CER and degradation (between clean and noisy transcripts) compared to word2vec.
The proposed confusion2vec gives the least CER among all the models (a relative improvement of 20.84\% over word2vec, 21.9\% over GloVe and 40.65\% over random initialization).
Moreover, C2V displays higher robustness going from clean to noisy ASR transcriptions (degradation, $\Delta_{\text{diff}}$, is minimal). A relative improvement in robustness of 37.48\%, 27\% and 51.36\% compared to GloVe, Word2vec and random initialization respectively is observed with C2V (in terms of $\Delta_{\text{diff}}$).
The confusion2vec word feature representation is able to use the embedded acoustic similarity information to recover from errors resulting from acoustically confusable words in the ASR output transcriptions.
\stepcounter{footnote}
\footnotetext{\label{footnote_exp_set}Original settings of \cite{liu2016joint,Liu+2016,li2018self}, make use of train + dev data for training. They pre-process data by substituting the digits with a token.}

Further, we compare our proposed system with the recent state-of-the-art works on SLU (see bottom part of Table~\ref{tab:results}).
Note, the recent works employ much more complex modeling techniques compared to ours.
Thus, as expected the recent works outperform our simple RNN architecture testing on clean transcriptions.
However, on noisy ASR transcriptions, even with a much simpler model, the proposed confusion2vec achieves significantly lower CER (a relative improvement of at-least 20.84\%) compared to state-of-the-art models.
Moreover, again, the degradation with confusion2vec is the least among all the models, a relative 37.48\% lesser degradation compared to the recent works.
The results highlight the potent robustness of the confusion2vec word feature representation.
In addition, we believe that the gains from the complex classification modeling are orthogonal to gains from confusion2vec word feature representations and thus should result in additional gains incorporating more complex models with confusion2vec.

\subsection{Training on ASR}
Further, we also perform additional experiments by training the intent classification models on noisy ASR transcripts.
A more robust feature representation should theoretically help in reducing the noise in the model translating to better performance.
From Table~\ref{tab:asr_train_results}, it is evident that all the models improve with the matched noisy train and test conditions.
The proposed confusion2vec based model gives the least CER among all the models.
The confusion2vec feature representation is better able to explain the (acoustic) errors and doing so reduces confusion and noise in the intent classification model, thereby resulting in a better and robust performance.
Moreover, comparing it with the recent study by \citet{schumann2018incorporating}, although the results are not directly comparable due to differences in the WER of the ASRs, our proposed method achieves a lower CER in-spite of much worse WER conditions\footnote{\label{footnote}We don't domain-constrain, optimize or rescore our ASR, as in \cite{schumann2018incorporating, liu2016joint}.  We treat ASR as an independent module for fair comparison with other models and for domain-generalization and portability of our system and conclusions.}.
This suggests that explicitly modeling in-domain ASR errors as in \cite{schumann2018incorporating} is of lesser effect compared to modeling the general acoustic signatures between words in a language as in the case with confusion2vec.

\begin{table}[t]
  \centering
  \begin{tabularx}{\columnwidth}{c *{2}{Y}}
    \toprule
     \textbf{Model}   & \textbf{WER \%} &\textbf{CER \%}  \\
    \midrule
    Random                   & 18.54   & 5.15 \\
    GloVe \cite{pennington2014glove}        & 18.54   & 6.94 \\
    Word2Vec \cite{mikolov2013distributed}  & 18.54   & 5.49 \\
    C2V (proposed)                & 18.54   & \textbf{4.70} \\
    \midrule
    \citet{schumann2018incorporating}      & 10.55   & 5.04\footnotemark[\getrefnumber{footnote}]  \\
    \bottomrule
  \end{tabularx}
  \captionsetup{justification=centering}
  \caption{Results with Training and Testing on ASR transcripts.}\label{tab:asr_train_results}
  \vspace{-8mm}
\end{table}

Finally, comparing the results from Table~\ref{tab:results} and Table~\ref{tab:asr_train_results}, it is encouraging to see that the proposed confusion2vec model trained on clean transcripts is able to inherit enough robustness to achieve lower CER (than GloVe) and comparable CER to the models (word2vec) trained on ASR output, possibly reducing the need for adaptation on ASR and allowing for more generalizable systems.

\section{Conclusion and Future Work}\label{sec:conclusion}
\heading{Conclusion}
In this paper, we proposed an intent detection model based on confusion2vec word vector representation targeting noisy ASR transcriptions.
The proposed word embeddings significantly outperform the popular leading word vector representations like word2vec and GloVe in the cases of noisy ASR output.
Comparisons are made with various recent state-of-the-art studies, and we find the proposed method improves over them by a considerable margin despite using relatively simple RNN architectures for classification.
The robustness of confusion2vec also extends to models trained on noisy ASR, achieving the least CER among the conventional word embedding as well as the recent studies.
Encouraging results suggest confusion2vec robustness to errors eliminates the need for adapting the intent classification models on noisy ASR outputs.

\heading{Future Work}
In future, we plan to apply and evaluate the proposed confusion2vec on additional SLU tasks like slot-filling, domain classification and named entity recognition.
We believe the proposed model should provide similar advantages, especially under noisy conditions.
Addressing multiple SLU tasks also allows us to use more complex joint-modelling systems with confusion2vec.
The better, more complex, models should provide improvements orthogonal to confusion2vec feature representations, and we thus expect to see further improvements.
We also plan to conduct more in-depth analysis on how the signal conditions and ASR performance affect each  model; we expect  confusion2vec to provide more gains as the ASR performance deteriorates.
Representation of multiple path outputs from the ASR with confusion2vec instead of only the best path is also a possible future direction.

\sevenpt
\bibliographystyle{IEEEtranN}
\bibliography{mybib}

\end{document}